# The signal is the ceiling

Measurement limits of LLM-predicted experience ratings from open-ended survey text

Andrew Hong, Jason Potteiger, Luis E. Zapata
*Dimension Labs*

April 2026

**Abstract**

An earlier paper (Hong, Potteiger, and Zapata 2026) established that an unoptimized GPT 4.1 prompt predicts fan-reported experience ratings within one point 67% of the time from open-ended survey text. This paper tests the relative impact of prompt design and model selection on that performance. We compared four configurations on approximately 10,000 post-game surveys from five MLB teams: the original baseline prompt and a moderately customized version, crossed with three GPT models (4.1, 4.1-mini, 5.2). Prompt customization added roughly two percentage points of within-±1 agreement on GPT 4.1 (from 67% to 69%). Both model swaps from that best configuration degraded performance: GPT 5.2 returned to the baseline, and GPT 4.1-mini fell six percentage points below it. Both levers combined were dwarfed by the input itself: across capable configurations, accuracy varied more than an order of magnitude more by the linguistic character of the text than by the choice of prompt or model. The ceiling has two parts. One is a bias in how the model reads text, which prompt design can correct. The other is a difference between what fans write about and what they actually decide, which no engineering can close because the missing information is not in the text. Prompt customization moved the first part; model selection moved neither reliably. The result is not that "prompt engineering helps a little" but that prompt engineering helps in a specific and predictable way, on the part of the ceiling it can reach.

## 1. Introduction

Post-game fan surveys collect two kinds of information from the same person about the same experience. The closed-ended question asks for a number: "On a scale from 0 to 10, how would you rate your overall experience?" The open-ended question asks for a written explanation of the rating given: what the fan noticed, what went well, what fell short. The number is a verdict, a single integer that compresses everything the fan experienced into a holistic judgment. The language is a record, a selective account of whichever moments were salient enough to mention. These two outputs are related, but they are not the same thing, and the relationship between them is the subject of this paper. An earlier paper on MLB fan surveys showed that an unoptimized GPT 4.1 prompt recovers the number within one point 67% of the time and argued that the residual gap is a construct difference rather than measurement error (Hong, Potteiger, and Zapata 2026). That paper positioned its baseline as a floor awaiting engineering. This paper tests what engineering can do to the floor: how much prompt design and model selection can actually change alignment, and in what direction?

We tested four annotation configurations on approximately 10,000 post-game fan surveys collected from five MLB teams between March and June 2025. Each configuration used a GPT model to read a single fan's open-ended text and predict the fan's self-reported overall experience rating on a 0 to 10 scale, receiving only the text, not the rating or any other information. The four configurations varied the prompt function (a generic baseline vs. a customized version) and the model (GPT 4.1, GPT 4.1-mini, and GPT 5.2). Predicting a number on a continuous scale from open-ended text is a different measurement task from the categorical classification that dominates the LLM annotation literature. The model must integrate multiple textual signals, positive and negative, specific and general, explicit and implied, into a single numeric estimate, and the error structure (not just the agreement rate) determines whether the measurement is fit for use.

Prompt design was the strongest engineering lever. The customized prompt on GPT 4.1 achieved 39% exact match and 69% within-±1 agreement, outperforming the generic baseline on every metric. The same customized prompt on GPT 4.1-mini produced the lowest alignment with survey scores of the four configurations, with exact match dropping to 20%, confirming that prompt and model must be validated together. But the largest source of variation in agreement was not the prompt or the model. It was the linguistic character of the input text. On unambiguously positive text, all capable models converged between 83% and 86% within ±1. On text containing negative operational details, alignment with the survey score dropped substantially, within-±1 agreement fell to roughly 42–44% across all three capable configurations (excludes 4.1-mini), consistent with a documented LLM measurement bias we address in Section 7. The customized prompt added about 3 percentage points on positive-only and no-lexicon-hit text, where the verdict is typically high and negativity bias has room to pull the prediction down, and about 2 points on negative-only text, where the surface signal and the verdict agree. This asymmetry is the signature of a bias-correction intervention, not a general reasoning improvement. The text itself was a far more powerful determinant of accuracy than any engineering choice: the gap between the best and worst text types was more than twenty times the gap between the best and worst capable configurations within any single text type. The text determines the ceiling; the prompt and model determine how efficiently the instrument reaches it.

Our previous paper[1] documented the pattern on this data: misalignment is systematic, not random, with 86% of large prediction errors being under-predictions concentrated where fans described operational friction (concession lines, parking, exits) but rated the overall experience positively. We argued this gap is not random noise around a correct answer but a structured divergence between two related but distinct cognitive outputs, where open-ended prompts activate retrieval of salient incidents while rating scales activate holistic evaluative judgments shaped by affect and the peak-end heuristic (Kahneman, Wakker, and Sarin 1997; Schwarz 1999). The present paper accepts that construct account and decomposes[2] the ceiling it implies. The ceiling has two components, and they respond to engineering differently. The cognitive component is the construct difference itself, invariant to prompt and model because the missing information is not in the text. The measurement component is an instrument bias correctable through prompt design that targets it specifically. The decomposition is what makes the engineering result interpretable rather than merely empirical: prompt customization gained roughly two percentage points because it intervened on the corrigible component, while model selection failed to transfer because neither swap reliably moved either component. The gap the prior paper documented persists, and it is the gap itself, not its closure, that carries the operational signal. The directional error structure that limits the predicted rating as a point estimate makes it a high-precision classifier for identifying top-tier and at-risk fans even at 39% exact match. Its value is not in matching the survey score but in diverging from it predictably.

---

[1] Hong, A., Potteiger, J., & Zapata, I. (2026). "LLM Predictive Scoring and Validation: Inferring Experience Ratings from Unstructured Text." arXiv:2604.14321.
[2] By "decomposition" we mean separating the ceiling into components that have different causes and respond differently to engineering intervention.

## 2. Related research

### 2.1 From annotation accuracy to measurement validity

The question of whether large language models can annotate text reliably has been answered; the question of what their annotations measure has not. The capability evidence is broad and convergent: zero-shot LLMs match or exceed crowd-sourced human annotation across political content (Gilardi et al. 2023; Törnberg 2025), multilingual psychological text (Rathje et al. 2024), and standard classification benchmarks. No serious dispute remains about whether LLMs can assign labels to text at scale. The dispute has moved to what the labels mean.

Ludwig, Mullainathan, and Rambachan (2024) argue that LLM outputs are best understood as measurements of latent constructs, subject to the full apparatus of measurement error and construct validity that any instrument requires. Once the field adopted this framing, a series of more specific vulnerabilities came into focus. Codebook definitions that deviate from a model's pretraining priors can cause zero-shot annotations to measure background concepts rather than researcher-specified constructs (Halterman and Keith 2024). High intercoder reliability does not guarantee valid measurement, because coders can agree consistently on the wrong thing (Kjell et al. 2025). And the choice of model is itself a consequential researcher degree of freedom: across fourteen published studies, different LLMs agreed with each other only moderately (pairwise Krippendorff's α of 0.16–0.69), and model choice frequently shifted downstream coefficient estimates, sometimes reversing statistical significance (Yang et al. 2025). A parallel line of work (Egami et al. 2023) addresses the same problem from the inference side, treating LLM-derived labels as imperfect surrogates whose noise must be modeled in downstream estimation rather than reduced at the instrument. The common thread is that annotation agreement is necessary but not sufficient. What matters is whether the labels capture the intended construct, whether the measurement is stable across reasonable implementation choices, and whether the specific configuration has been validated against ground truth.

One finding from this literature bears directly on a design choice in the present study. Asirvatham, Mokski, and Shleifer (2025) tested 100 dramatically varied prompt wordings (ranging from 32-word telegraphic instructions to 563-word Shakespearean prose) on the same attribute-rating task and found that prompt wrapper variation had minimal effect on measurement correlations. Radical differences in tone, length, and phrasing barely moved the output. But their variation was in the wrapper text surrounding the measurement task, not in the task's cognitive architecture. The customized prompt tested in the present study does not change the wrapper. It adds a three-step decision process (prioritize an explicit verdict, then weigh positives against negatives, then do not over-penalize isolated complaints) and a rationale field requiring the model to explain its scoring logic[3]. These are structural interventions in how the model reasons about the evidence, not cosmetic variations in how the task is described. If wrapper text does not affect measurement but decision architecture does, the design lever for practitioners is narrower and more specific than the general injunction to "write better prompts." Carlson et al. (2026) arrive at a compatible conclusion from a different direction: subtle implementation choices (prompt design, model selection, temperature) can significantly alter downstream research conclusions, and systematic sensitivity analysis across configurations is the appropriate response.

One specific use of decision-architecture-level prompt design is bias correction. LLMs pretrained on large text corpora inherit statistical associations from their training data. Most relevant here, an association between surface-negative lexicon and low scalar ratings, a pattern well-documented in product review, restaurant review, and app store review data. This association is functional in domains where negative language and low ratings genuinely covary, but it produces systematic over-penalization in domains where the two decouple. Structured prompt instructions that explicitly weigh overall verdict against surface valence are one mechanism for counteracting this tendency, and we argue in Section 4 and Section 7 that this is a principal function of the customized prompt tested here.

## 2.2 From categorical classification to scalar measurement

Nearly all of the validation work reviewed above was conducted on categorical classification: sentiment labels, topic codes, political stance, named entities. These tasks share a structure that makes evaluation tractable: a fixed label set, a human gold standard, and a codebook that defines what each label means. The problems of LLM-based measurement become qualitatively different when the output is not a category but a number on a continuous scale.

Licht et al. (2025) document what goes wrong. Evaluating LLMs on scalar construct measurement across multiple political science datasets, they find that pointwise scoring suffers from systematic bunching: models cluster their outputs around a few preferred values rather than using the full scale, and the resulting distributions do not align with reference scores obtained from human pairwise judgments. The miscalibration is not a failure of any particular model or prompt. It is a property of asking LLMs to produce numeric ratings directly. The scale itself becomes a source of measurement error, because the model's relationship to the numbers it outputs is not the same as a human rater's relationship to the same scale.

This matters because when the output is a number rather than a label, the error structure carries information that categorical agreement metrics cannot capture. A model that assigns a 6 when the human assigned a 7 has made a different kind of "mistake" than one that assigns a 3, and the direction of the miss may be as informative as the magnitude. Kucia (2026) provides evidence that the direction is systematic. Evaluating LLMs as essay scorers under holistic and analytic rubrics, he documents large and stable negative directional bias. Kucia's finding is specific in an important way: the under-scoring persists across prompt variations and model choices, suggesting it is a property of how LLMs map text to scalar ratings rather than an artifact of any particular implementation. This is the baseline tendency any prompt architecture in this study is operating against. A customized prompt that explicitly instructs the model to avoid over-penalizing isolated complaints is intervening on a documented bias, not a hypothetical one.

---

[3] The rationale field is a post-hoc annotation in both configurations: models produce plausible justifications that do not reflect the actual features driving the prediction. The generated text is a narration the model produces; the computation happens in the forward pass and is not introspectively accessible (Turpin et al. 2023). The impact of field order is a subject for future study.

Predicting a self-reported experience rating from open-ended feedback is a distinct instance of the scalar measurement problem. The input is shorter and less structured than essays or political speeches, the target is a respondent's own holistic verdict rather than a score against an external rubric, and the construct itself is whatever each respondent decided "overall experience" meant to them. The model must integrate multiple textual signals (positive and negative, specific and general, explicit and implied) into a single numeric estimate. The evaluation requires not just agreement metrics but an account of the error structure: whether the errors are symmetric or directional, whether they concentrate at particular points on the scale, and whether identifiable properties of the input text predict where the measurement succeeds and where it fails. This study tests that task directly, and the pattern it finds is consistent with the scalar measurement challenges Licht et al. and Kucia document in other domains: systematic under-prediction, concentrated in the midrange, tied to specific textual features.

## 2.3 Why the text and the number should diverge

The interpretive question this study raises, why fans who describe operational friction still rate the overall experience positively, has a specific theoretical prediction outside the LLM measurement literature. If the open-ended text and the numerical rating are produced by different cognitive processes, they should diverge in predictable ways, and the divergence should be directional rather than random.

The psychological literature offers three mechanisms that all point in the same direction. Kahneman, Wakker, and Sarin (1997) draw a formal distinction between experienced utility (the hedonic quality of an experience as it unfolds, moment by moment) and remembered utility (the retrospective evaluation of a past episode). Remembered utility is systematically biased: it follows the peak-end rule, weighting the most intense moment and the final moment while neglecting duration and the accumulation of lesser events (Kahneman et al. 1993). An open-ended survey prompt and a rating scale activate different sides of this distinction. Schwarz (1999) provides the measurement-side mechanism: open-ended prompts trigger retrieval of the most cognitively accessible specific incidents (concrete, vivid, easy to articulate), while closed-ended rating scales activate a holistic evaluative summary shaped by the peak-end heuristic and by current affect. Schwarz and Clore (1983) established the deeper reason: when people make global evaluative judgments, they use their affective state as a direct informational input, asking themselves "how do I feel about this?" rather than performing a deliberate aggregation of specific incidents. A fan whose team won an exciting game carries residual positive affect that serves as the primary input to the holistic rating, even if the specific incidents they recall and report include long concession lines and a confusing exit.

The prediction that follows is specific. The open-ended text should be dominated by salient operational details (what went wrong, what stood out) because these are the incidents most cognitively available for verbal report. The overall rating should reflect an integrative judgment that discounts those details in favor of emotional peaks and endings. An LLM reading only the text has access to the first process but not the second. It sees what the fan chose to describe, not how the fan chose to weigh it, producing exactly the directional under-prediction this study finds.

This is not the only possible explanation. The model may carry a negativity bias from pretraining: exposure to review corpora in which negative language correlates with low scores may cause it to over-penalize negative details regardless of context. And the survey score itself may be inflated by positivity bias: social desirability, consistency motivation, or the tendency to rate experiences at live events generously because attendance itself signals endorsement. The present data cannot fully adjudicate between these mechanisms. What they can do is test whether the divergence pattern is consistent with the cognitive prediction: systematic, directional, and concentrated where mixed signals are most common.

That pattern has been established on the current dataset (Hong et al., "Predictive Ratings," 2026). This paper accepts the construct-difference starting point and asks what engineering can do against the ceiling it implies.

## 3. Data, annotation, and validation design

### 3.1 Survey instrument and sample

The data come from post-game fan surveys administered by five Major League Baseball teams during the 2025 season, covering games played between March 28 and June 23, 2025. Each survey asked the fan a closed-ended rating question and one or more open-ended follow-up questions about their experience at the ballpark. The closed-ended question, overall_numrat—asked: "On a scale from 0 to 10, how would you rate your overall experience at the [specific] game on [date]?" This self-reported score serves as the ground truth for the validation.

The combined dataset contains approximately 10,000 completed surveys. The distribution of overall_numrat is positively skewed, as is typical of post-experience satisfaction surveys: roughly half of respondents rated their experience a 9 or 10, with progressively smaller counts at each lower score level. This skew is not an artifact of the collection method; it reflects the empirical reality that most fans who attend a baseball game and complete a post-game survey report a positive experience. The skew does, however, shape the validation in important ways: agreement metrics are dominated by the high end of the scale, and the midrange scores (5–8), though fewer in number, represent the cases where the measurement is most challenged.

The open-ended text responses vary substantially in length, specificity, and linguistic clarity. Some fans wrote a single phrase ("great time"); others produced multi-paragraph narratives describing concession lines, weather, seat quality, and interactions with staff. The predicted rating was derived from these open-ended responses—the model received only the text, not the numerical score, and was asked to infer what the fan would have rated.

## 3.2 Annotation methodology

The annotation was executed on the Dimension Labs language data platform using a prompt function, a JSON schema that defines the large language model's task and output structure (Appendix A). Each open-end response was processed independently: the model received one fan's text as input and returned a structured set of labeled fields for that response alone. This session-level independence ensures that each predicted rating reflects only the language in a single fan's text, making the resulting scores comparable across responses, teams, and time periods.

We tested four annotation configurations in a two-factor design: two prompt functions crossed with three GPT models. The **baseline prompt function** (p2_1), originally introduced and validated by Hong, Potteiger, and Zapata (2026), used a generic instruction to infer the fan's overall experience rating on a 0–10 scale, producing three fields: predicted_rating (integer), predicted_rating_confidence_score (High, Medium, or Low), and predicted_rating_evidence (a brief text summary of the strongest evidence). The **customized prompt function** (p3_1_*) added two design features informed by early error analysis: a three-step decision process (prioritize an explicit overall verdict if present, then weigh positives and negatives, then do not over-penalize minor complaints if the overall tone is positive) and a predicted_rating_rationale field requiring the model to explain its scoring logic. Both prompt functions framed the task identically: estimate the fan's self-reported overall experience rating as they would answer the survey question.

The baseline was run on the full dataset using GPT 4.1; the customized prompt was run on the full dataset using three models: GPT 4.1, GPT 4.1-mini, and GPT 5.2. All four configurations cover the same collection window and are evaluated on the joint sample of 9,961 sessions present across every run, enabling direct prompt-vs-model comparisons on the same population. To assess prompt-level reliability independent of model variation, the baseline configuration was additionally run twice more on the same data (p2_2, p2_3); these independent re-runs are reported in Appendix B Tables B3, B4, B5, and B10 and underpin the prompt-stability statistics referenced in Section 8.

**Table 1. Attributes used in the validation, with source and operational definition.**

| Attribute | Source | Definition |
|---|---|---|
| overall_numrat | Self-reported | Fan's overall experience rating, 0–10 |
| predicted_rating | LLM-annotated | Model's estimate of overall rating, 0–10 integer, from text only |
| confidence_score | LLM-annotated | Evidence clarity: High, Medium, or Low |
| rationale | LLM-annotated | Model's explanation of scoring logic (customized prompt only) |
| evidence | LLM-annotated | Brief summary of strongest textual evidence |

### 3.3 Validation metrics

No single agreement metric can characterize a measurement that fails differently at different points on the scale: strict agreement, tolerance for near-misses, average magnitude, and directional bias each reveal a different aspect of how the predicted rating relates to the survey score. We evaluate alignment using four complementary metrics. **Exact match** is the proportion of sessions where the predicted rating equals the survey rating. This is the strictest test: it asks whether the model can recover the precise integer the fan chose. On a 0–10 scale, random guessing would produce exact match rates near 9%, so any rate substantially above this baseline indicates real signal. **Within ±1** is the proportion of sessions where the predicted rating falls within one point of the survey rating. This is the more practically relevant metric: if downstream use can tolerate a one-point margin, within-±1 agreement is the standard that matters. **Mean absolute error (MAE)** captures the average magnitude of the miss across all sessions. An MAE of 1.0 means the model is, on average, one point away from the fan's score. **Mean signed error** captures directional bias, specifically whether the model systematically over-predicts or under-predicts. A negative mean signed error indicates systematic under-prediction.

All rates are reported with 95% confidence intervals using normal approximation margins of error. Throughout the paper, we refer to three score bands for interpretive convenience: **low** (survey scores 0–4), **moderate** (5–8), and **high** (9–10). These are descriptive groupings chosen to reflect the empirical patterns in the data.

## 4. Prompt customization modestly improved agreement; not model

Consider a fan who wrote: "It was a good game, and I can't complain about the product on the field […]." This fan rated the overall experience a 9; the model predicted 8, a one-point miss that captures the overall verdict while slightly discounting the measured praise. How the model weighs mild qualification against overall tone is a function of what the prompt tells it to do. We tested whether explicit prompt instructions to prioritize the overall verdict and not over-penalize isolated complaints would improve alignment with the fan's self-reported score, and whether the choice of model would amplify or undermine that improvement.

The customized prompt on GPT 4.1 (`p3_1_41`) produced the strongest alignment of any configuration tested. It achieved 39% exact match, 69% within ±1, and a mean absolute error of 1.22, outperforming the generic baseline on every metric (36% exact, 67% within ±1, MAE 1.29). The improvement is consistent across the score distribution and statistically significant given the sample sizes (the 95% confidence intervals for exact match do not overlap: 39.4% ± 1.0pp vs. 36.3% ± 0.9pp). The customized prompt's three-step decision process and the addition of the rationale field appear to be doing real work: by requiring the model to articulate why it assigned a score, the prompt forces a more deliberate weighing of the evidence rather than a reflexive response to surface-level tone.

The improvement is concentrated in text regimes where surface valence and the fan's verdict diverge (positively-valenced text, mixed text, and neutrally-worded text), consistent with the customized prompt functioning as a bias-correction intervention rather than a general agreement lift. Section 7 develops this interpretation using the breakdown by lexical clarity.

The complication arrives with the model comparison. One might expect that running the same well-engineered prompt on a different model would produce comparable or better results: that the prompt carries the intelligence and the model provides the capability. The data in this test do not support this. Unsurprisingly, GPT 4.1-mini had worse precision; it produced the worst alignment of all four configurations (20% exact match, 61% within ±1, MAE 1.51). The drop is not marginal: exact match fell by roughly 20 percentage points and the gap between predicted and actual scores nearly doubled. The mini model under-scored 70% of all sessions, as if it could not hold the distinction between "the fan mentioned something negative" and "the fan's overall experience was negative." Somewhat surprisingly, GPT 5.2 fell between the baseline and the customized GPT 4.1 configuration, trailing the customized GPT 4.1 on every metric despite being a more advanced model.

Table 2. Overall alignment metrics across four prompt-and-model configurations.

| Run | Prompt | Model | N | Exact | Within ±1 | MAE | Signed err. |
|---|---|---|---|---|---|---|---|
| p2_1 | Baseline | GPT 4.1 | 9,961 | 36% | 67% | 1.29 | −0.96 |
| p3_1_41 | Custom | GPT 4.1 | 9,961 | 39% | 69% | 1.22 | −0.83 |
| p3_1_41mini | Custom | 4.1-mini | 9,961 | 20% | 61% | 1.51 | −1.18 |
| p3_1_52 | Custom | GPT 5.2 | 9,961 | 32% | 67% | 1.38 | −1.09 |

The error directionality sharpens the picture. All four runs systematically under-predict: the model assigns lower ratings than the fan chose, with mean signed errors ranging from −0.83 (the best run) to −1.18 (the mini). The under-prediction is not a property of one configuration. It is a structural feature of the task.

**Figure 1. Alignment across configurations.**

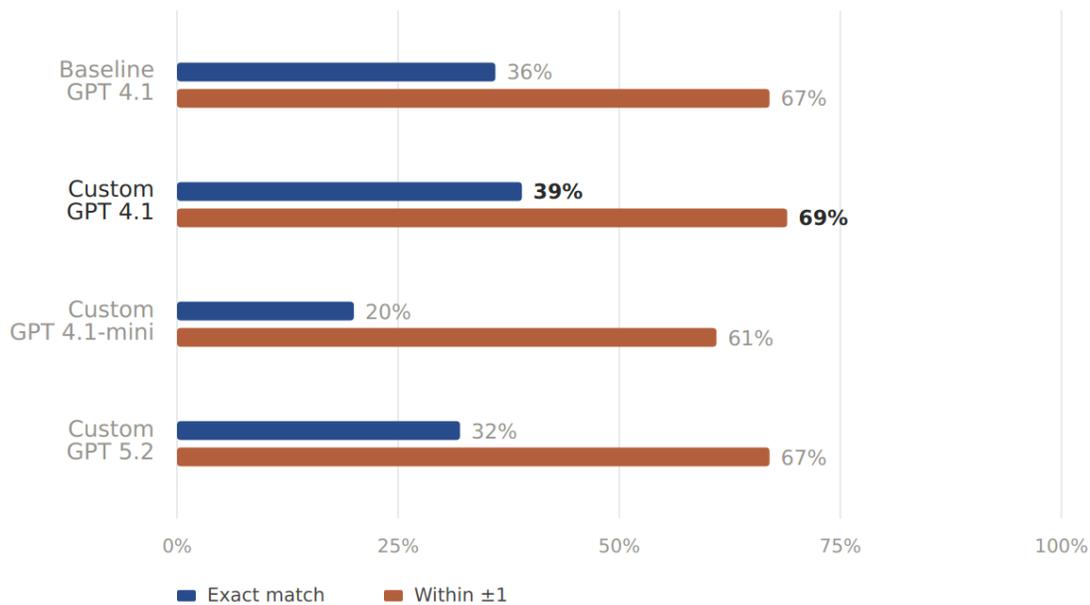

**Prompt engineering improves alignment**

Exact-match and within-±1 agreement rates across four prompt–model configurations, comparing baseline and customized prompts on three model variants.

*Each row shows one prompt–model configuration. Compare the two colored bars to see how exact match and within-±1 rates differ across runs.*

overall_alignment_with_moe · exact match and within ±1 · n=9,961 · Figure 1

**Figure 2. Score distribution by model.**

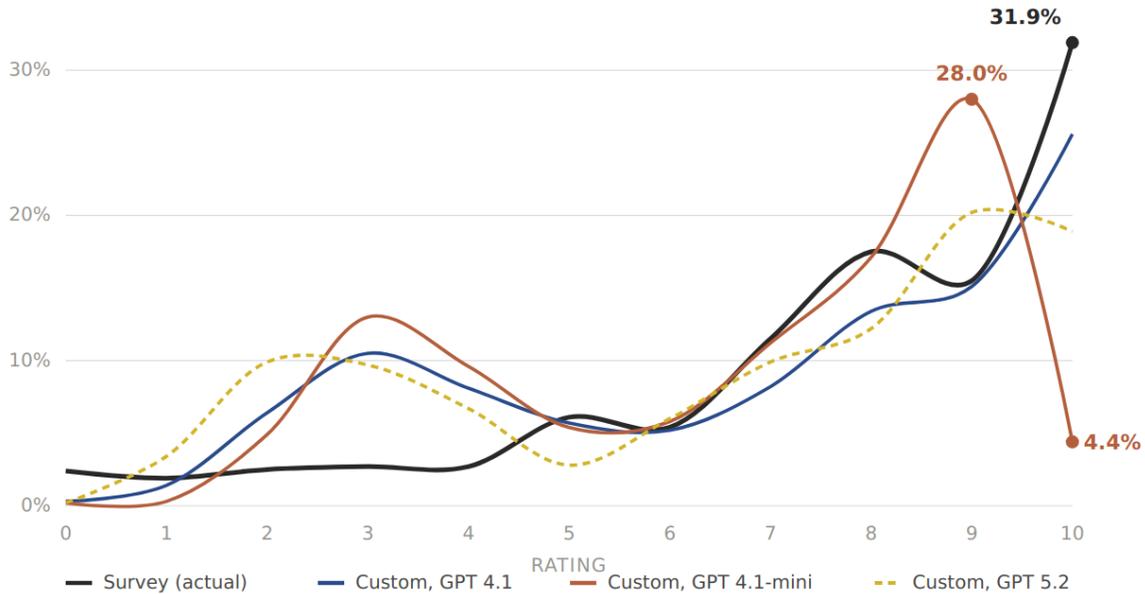

**Model choice impacts low and high predictions most**

Predicted score distributions from three model configurations compared to the actual survey response distribution, across rating levels 0–10.

*Each line traces how often a model predicts each rating level. Where a colored line diverges from the black survey line, the model is over- or under-predicting that score.*

marginal predicted-score distributions · survey distribution from consistency_by_score_level · n=9,961 · Figure 2

The practical implication is that prompt engineering and model selection must be validated together. A prompt optimized for one model does not transfer reliably to another, and the failure mode is not degraded precision but altered behavior.

The four configurations divide into two groups. The three capable configurations (the baseline, the customized prompt on GPT 4.1, and the customized prompt on GPT 5.2) produce within-±1 agreement rates of 67.1%, 69.3%, and 67.1% respectively, a spread of roughly two percentage points. GPT 4.1-mini sits apart: at 61.4% within ±1 with a predicted-score distribution qualitatively unlike the other three (Figure 2), it is not simply a less accurate instrument but a differently behaved one. **In the remainder of the paper, 'capable configurations' refers to the three non-mini runs;** claims about alignment variation across prompt and model choices exclude mini unless noted.

# 5. Alignment varies by score level, and the pattern is consistent across configurations

Consider two fans. One rated the experience a 10 and wrote "amazing night, everything was perfect." The other rated it a 7 and wrote about a great game undermined by slow concessions and a confusing exit. The model's task in the first case is pattern recognition; in the second, it is an act of judgment about how competing signals resolve into a single number. The text does not record how the fan resolved them.

The score-level breakdown confirms that this difference in task difficulty, not prompt or model quality, drives the alignment pattern. Where the text and the verdict agree, the model succeeds easily: at survey score 10, within-±1 agreement reaches 93% and MAE drops to 0.42. Where the text contains mixed signals that the fan's verdict resolved but did not record, the model struggles: at survey score 6, within-±1 agreement falls to 34% and MAE rises to 2.05. The consistency across all four configurations is the finding. Every run hits the same wall at the same score levels. The midrange challenge is a property of where the verdict and the textual signal are most likely to diverge (fans with good days who wrote about specific friction), not a deficiency of any particular configuration.

A single case illustrates the nature of the difficulty. One fan wrote: "The game was exciting and we won!! BUT, I did have a VERY unpleasant experience [...]" with a vendor. This fan rated the overall experience an 8; the model predicted 4. The text contains both an enthusiastic verdict and a vivid negative incident. The model weighted the incident more heavily than the fan did. No amount of prompt engineering can tell the model how much weight this fan placed on the vendor interaction relative to the game itself, because that information is not in the text. It is in the number the fan chose.

**Table 3. Alignment by self-reported survey score level for the best run (p3_1_41).**

| Survey score | N | Exact match | Within ±1 | MAE |
|---|---|---|---|---|
| 0 | 236 | 12% | 32% | 2.00 |
| 1 | 186 | 15% | 62% | 1.45 |
| 2 | 250 | 37% | 80% | 0.91 |
| 3 | 268 | 37% | 85% | 0.84 |
| 4 | 268 | 24% | 68% | 1.13 |
| 5 | 607 | 14% | 42% | 1.67 |
| 6 | 542 | 12% | 34% | 2.05 |
| 7 | 1,145 | 23% | 47% | 1.95 |
| 8 | 1,752 | 30% | 60% | 1.66 |
| 9 | 1,554 | 34% | 72% | 1.33 |
| 10 | 3,191 | 68% | 93% | 0.42 |

What makes the midrange failures analytically tractable is their directionality. At survey scores 5 through 8, the errors are overwhelmingly under-predictions, not random scatter but a systematic pattern in which the model reads the text and arrives at a more negative assessment than the fan's overall verdict. The 30% of predictions that fall outside ±1 concentrate in the moderate band, they point in one direction, and they raise a specific question: why does the model penalize more than the fan does?

**Figure 3. Error composition by score.**

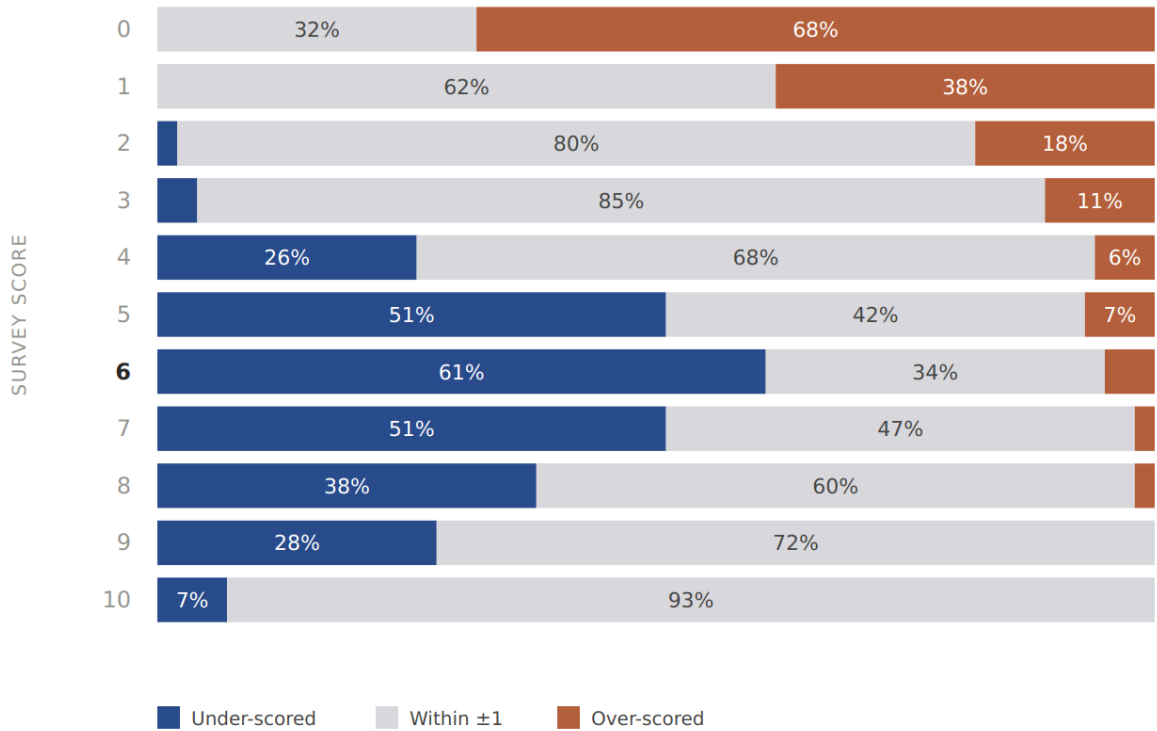

misalignment by survey rating level · p3_1_41 · n=9,961 · each row sums to 100% · Figure 3

## 6. Diagnosing the 30%: misalignment is systematic, concentrated, and traceable

Seven out of ten predicted ratings fall within one point of the fan's self-reported score. The other three are the subject of this section: not to explain them away but to determine whether roughly 3,000 sessions of disagreement are random noise or a structured signal that reveals what the predicted rating measures when it diverges from the survey score.

## 6.1 The errors are almost entirely in one direction, and severity amplifies the asymmetry

The errors are not scattered in both directions. Of the roughly 3,000 sessions outside ±1 in the best run, 86% are under-predictions: the model assigned a rating at least two points below the fan's self-reported score. The asymmetry intensifies with severity. Among three-point errors, 89% are under-predictions. At four points, 95%. At five or more, 96%. The model does not wander in both directions when it misses badly. It almost always misses downward.

The concentration of misalignment across the score range mirrors this directionality. At the extremes, the model has little trouble: at survey score 10, only 7% of predictions fall outside ±1, and at survey score 3, only 15%. These are cases where the text and the verdict are likely to agree. But at survey score 6, the misalignment rate jumps to 66%, and at survey score 7, to 53%. The model's own confidence labels track this pattern: medium-confidence predictions misalign at 57%, compared to 25% for high confidence. Yet even among high-confidence predictions, one in four falls outside ±1. These are cases where the model judged the text to be clear evidence for a particular rating, and the fan's verdict disagreed. The divergence is not about ambiguity alone but about what the model treats as evidence versus what the fan treats as decisive.

A seven-point miss illustrates how extreme the divergence can be. One fan rated the experience a 9 but opened their response by describing a single negative incident: "I selected this number for one distinct reason [...]." The model, reading a text dominated by a complaint narrative, predicted 2. The fan's overall verdict and the fan's text pointed in opposite directions, and the model followed the text.

**Table 4. Error magnitude distribution with directional breakdown (p3_1_41).**

| Absolute error | N | % Under-scored | % Over-scored |
| --- | --- | --- | --- |
| 0 (exact) | 3,936 | — | — |
| 1 | 2,998 | 72% | 28% |
| 2 | 1,366 | 78% | 22% |
| 3 | 840 | 89% | 11% |
| 4 | 507 | 95% | 5% |
| 5+ | 437 | 96% | 4% |

## 6.2 Operational friction narratives are the strongest correlates of misalignment

If the model is systematically under-scoring, what specific textual features are associated with the errors? We tested this using a set of lightweight keyword flags designed to identify common experience themes in the fan responses.1 These are correlational proxies, not causal labels. But the pattern of overrepresentation identifies candidate mechanisms worth investigating.

Transcripts mentioning lines, wait times, or queuing misalign at 39%, compared to 29% for transcripts without these terms, a lift of 1.35×. Parking, exit, or traffic references produce a similar pattern: 38% misalignment (1.27× lift). The compound effect is the most telling finding. When both sets of flags are present in the same transcript, the misalignment rate rises to 41% (1.46× lift) and the MAE to 1.49. When neither flag is present, only 28% misalign.

**Figure 4. Operational friction and misalignment.**

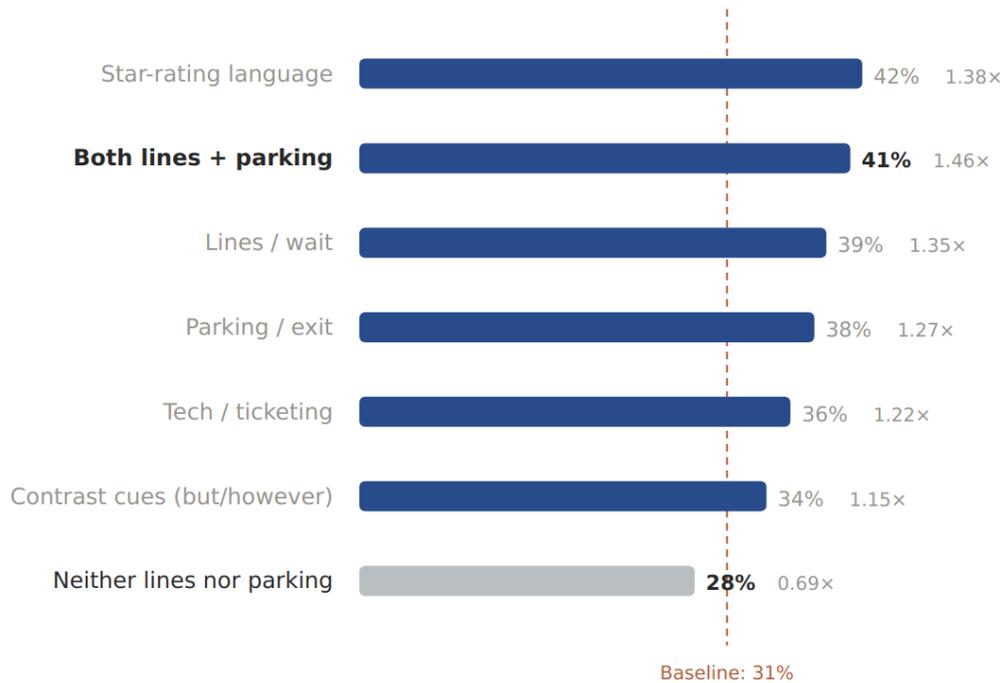

*Bars are sorted by misalignment rate. Keyword flags to the right of the dashed baseline indicate higher-than-average misalignment when that term appears in the transcript.*

reason-proxy mining · p3_1_41 · misalignment = outside ±1 · baseline 30.7% · n=9,961 · Figure 4

The cases in Table 6 illustrate how the mechanism operates in practice. A fan who writes sarcastically about concession lines and rates the experience a 10 has produced a text that points in the opposite direction from the verdict. A fan who opens with enthusiasm about the game and then describes a policy grievance has produced a text where the negative content is more vivid and specific than the positive framing. In each case, the model is not wrong about the text. The text does contain negative operational details. The model is wrong about how much those details should reduce the overall rating, because the fan's internal weighting is not in the text. It is in the number the fan chose. This is the pattern an LLM inheriting negativity bias from its training data would produce: specific, concrete, vivid negative details exert more weight in the model's prediction than in the fan's own evaluative integration.

Table 6. Divergence patterns in operational friction cases.

| Survey | Pred. | Pattern | Fan text (excerpt) |
|---|---|---|---|
| 10 | 3 | **Sarcastic friction** | "I appreciated waiting in the long concession lines. Loved the strain in my neck [...]" |
| 1 | 9 | **Policy grievance behind positive opening** | "We loved the game and all the professionalism with the rain delay [...] However we were told we could receive additional free tickets" |
| 10 | 2 | **Dual rating in text** | "It was due to my friends around me. A 10 but rating this exact game a 2 [...]" |

## 7. The text, not the prompt, determines the alignment ceiling

A data team choosing between prompt variants will find a spread of roughly two percentage points in within-±1 agreement across capable configurations. A team sorting the same predictions by the clarity of the input text will find a gap more than twenty times that size. The binding constraint on prediction agreement is not the prompt, the model, or the engineering: it is the linguistic character of the text itself. Prompt and model choices operate within the space the text allows, and the most defensible role for prompt design is not to close the gap between text and verdict, which it cannot do, but to correct measurement tendencies the instrument imports from pretraining. This section develops that argument across three dimensions: model confidence as a proxy for text-rating coherence, lexical clarity as the strongest single predictor of alignment, and the two mechanisms that jointly produce the ceiling.

### 7.1 Confidence labels separate reliable predictions from unreliable ones

The confidence label assigned by each model is the single most powerful predictor of whether a prediction will be accurate, more powerful than the choice of prompt, model, or any textual feature tested in isolation. But the label is misnamed. The model is not rating its internal certainty in the way a calibrated probability would. It is observing the text alongside the rating it has produced and judging how well the two cohere. The performance gap confirms this interpretation: in the best run (p3_1_41), high-confidence predictions achieve 75% within ±1 and MAE 1.06, while medium-confidence predictions drop to 45% within ±1 and MAE 1.86 (Figure 5; the analogous baseline figures appear in Table B6). The 30-percentage-point gap dwarfs the two-percentage-point spread across the three capable configurations (see Section 4 on the exclusion of GPT 4.1-mini as qualitatively different).

We use the term text-rating coherence throughout the remainder of this paper to describe what the confidence label actually captures: not "how sure the model is" but "how clearly the text supports a single rating." This reframing matters for practitioners because text-rating coherence is an observable property of the prediction that can be computed before ground truth is available. A data team can use it to gate which predictions enter downstream analytics and which require human review, without ever seeing the survey score.

**Figure 5. Confidence gating across runs.**

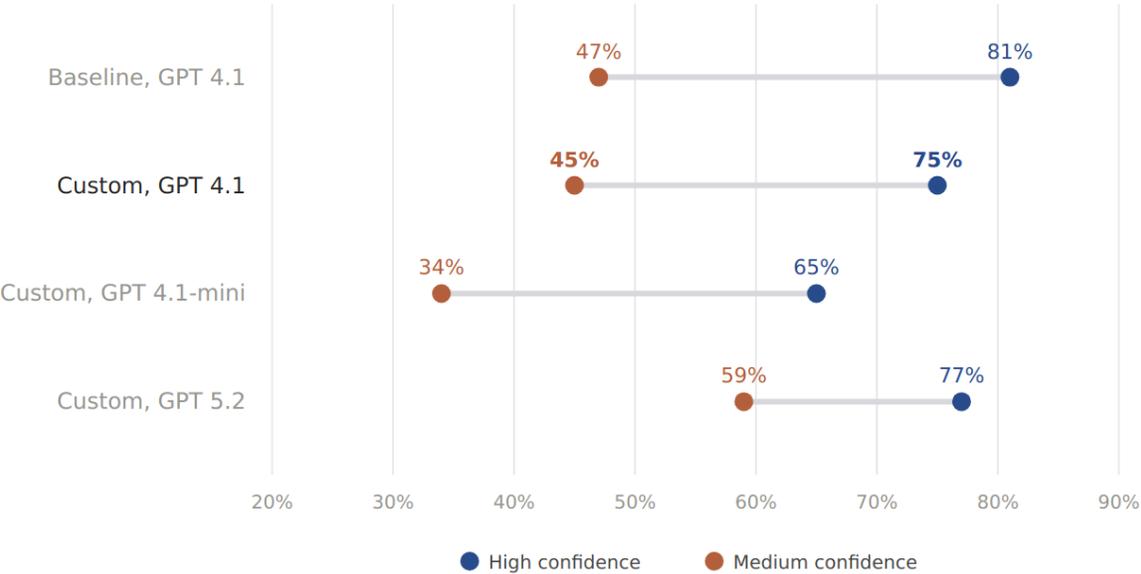

Each row is a prompt–model run. The gap between the two dots shows how much better high-confidence predictions perform compared to medium-confidence ones.

confidence_gating · within ±1 by confidence label · n=9,961 · Figure 5

The pattern across all of these features converges on a single mechanism. The textual elements most associated with misalignment are specific, concrete descriptions of negative operational experiences. The model reads the friction and assigns it weight; the fan's overall verdict absorbs the friction into a still-positive summary. The model is not wrong about the text: the text does contain negative operational details. It is wrong about how much those details should reduce the overall rating, because the fan's internal weighting is not in the text. It is in the number the fan chose.

## 7.2 Lexical clarity is the strongest single predictor of alignment, and the pattern reveals what the prompt is doing

The same instrument measuring the same construct produces radically different alignment with the survey score depending on a property of the input text that is observable before any external validation: whether the text contains clear positive language, clear negative language, or a mix. Table 8 reports alignment metrics for the best run (p3_1_41) by lexical clarity bucket.

Table 8. Alignment by lexicon clarity bucket for the best run (p3_1_41).

| Lexicon bucket | N | Exact match | Within ±1 | MAE |
| --- | --- | --- | --- | --- |
| Positive only | 4,239 | 53% | 86% | 0.69 |
| Mixed (both) | 767 | 33% | 68% | 1.20 |
| No lexicon hits | 4,189 | 31% | 58% | 1.58 |
| Negative only | 801 | 19% | 44% | 2.06 |

On positive-only text, within-±1 alignment reaches 86% (95% CI: 85–87%) with an MAE of 0.69. On negative-only text, the same run drops to 44% within ±1 (95% CI: 41–48%) with an MAE of 2.06. The 42-percentage-point gap is the single largest effect in the study, substantially larger than any effect of prompt or model choice within a single text type.

**Figure 6. Alignment by lexical clarity.**

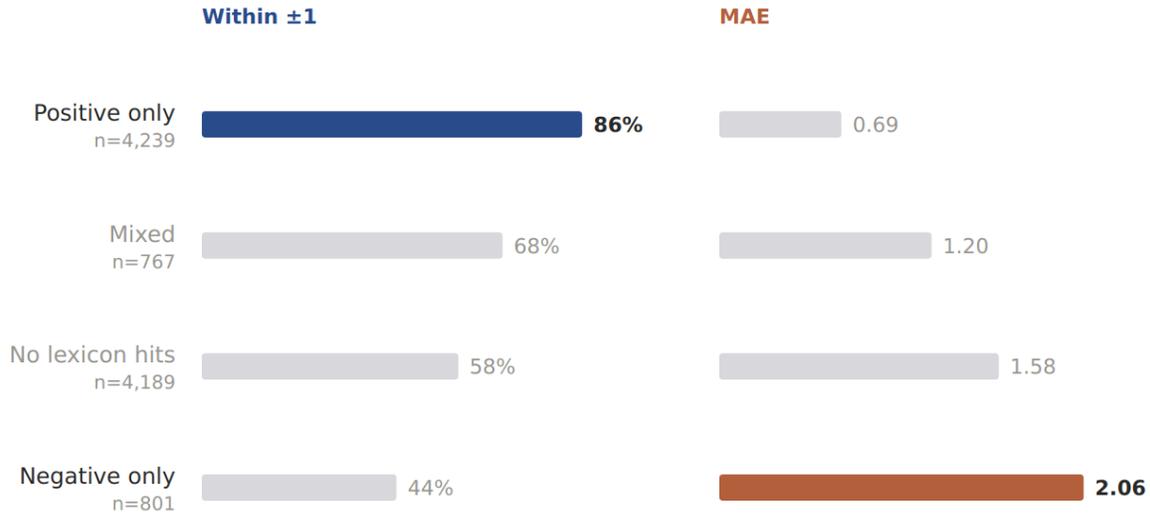

Left panel: higher bars indicate better agreement with survey scores. Right panel: lower bars indicate smaller prediction errors. Read across a row to see both metrics for the same bucket.

transcript_lexicon_bucket_metrics · p3_1_41 · n=9,961 · Figure 6

The "no lexicon hits" bucket (n=4,189) provides a large-sample midpoint: 58% within ±1, falling between the positive-only and negative-only extremes and consistent with the gradient. Texts with no clear sentiment keywords are precisely the cases where the model has no strong lexical signal to anchor its prediction, and their alignment falls accordingly.

**The customized prompt improves alignment across text types**. The customized prompt on GPT 4.1 outperforms the baseline in every lexicon bucket — positive-only (86% vs 83% within ±1), mixed (68% vs 68%), no-lexicon-hits (58% vs 55%), and negative-only (44% vs 42%). The improvement averages roughly two percentage points and is directionally consistent across all four text types. No regime exists in which the customized prompt underperforms the baseline.

This uniform pattern supports a specific interpretation of what the customized prompt is doing. LLMs trained on large text corpora (in particular on review data where negative language and low ratings genuinely covary) inherit a statistical association between surface-negative lexicon and low scalar ratings. Kucia (2026) documents this as a stable negative directional bias in LLM scoring that persists across prompt variations and model choices. Baumeister et al. (2001) establish the broader principle: negative information carries disproportionate weight in evaluative processing, a tendency that pretraining on human-generated text can only amplify in the model. The customized prompt's three-step decision process and rationale field are a bias-correction intervention against this tendency. They do not change what the text says; they change how the model weights negative content against overall tone when assigning a rating. The consistent direction of the improvement across all text regimes is the signature of an instrument being corrected for a known measurement bias rather than an instrument being improved by equal amounts on every reading. The improvement is largest on text where surface negativity has the most room to distort the verdict: roughly 3 percentage points on no-lexicon-hit text and positive-only text, versus 2 points on negative-only text where the verdict and the surface signal already point the same direction.

### 7.3 What the ceiling is made of

The ceiling has identifiable components, and they have different sources. This section catalogs them; Section 10 returns to what the asymmetry between them implies for engineering and deployment.

The diagnostic findings from Section 6 and the clarity analysis in this section converge on the mechanism predicted in Section 2.3. If the open-ended text and the survey score are outputs of different cognitive processes, the divergence should be systematic rather than random, directional rather than symmetric, and concentrated where mixed signals are most common. The data confirm each element. The divergence is systematic: 86% of large errors are under-predictions, not a scatter in both directions. It is concentrated: misalignment peaks in the moderate band (survey scores 5 through 8) where fans described friction they ultimately forgave. And it is tied to specific, identifiable content features: operational friction narratives (lines, wait times, parking, exit) are overrepresented among misaligned cases by 1.3 to 1.5 times baseline. The pattern is not what random measurement error looks like. It is what a structured construct difference looks like.

The text and the survey score are not two noisy measurements of the same thing. They are measurements of different cognitive outputs. The text records what was salient: the concrete, vivid, easily articulated details that the fan retrieved in response to an open-ended prompt. The survey score records a retrospective verdict shaped by emotional peaks, the ending of the experience, and residual affect (Kahneman, Wakker, and Sarin 1997; Schwarz 1999; Schwarz and Clore 1983). An LLM reading the text has access to the first process but not the second. It sees what the fan chose to describe, not how the fan chose to weigh it. The systematic under-prediction follows directly: the model assigns weight to negative details in proportion to their textual prominence, while the fan's verdict discounts them in favor of an integrative judgment the text does not contain. One fan illustrated this vividly by writing sarcastically about long concession lines and physical discomfort, then rating the overall experience a 10. The model, reading text dominated by negatively valenced language, predicted 3. The fan's overall verdict and the fan's textual account pointed in opposite directions, and the model followed the text.

The salience-integration divergence predicted by the cognitive science literature is not the only mechanism operating in the gap. An LLM reading the text also imports statistical priors from its pretraining—most consequentially, an association between surface-negative lexicon and low scalar ratings that is appropriate to the training domain but systematically over-weights negative description in this one. The customized prompt was designed to correct for this tendency, and the cross-text-type pattern documented in Section 7.2 is consistent with the correction succeeding: uniform improvement across positive, mixed, neutral, and negative text, with the size of the improvement largest where the bias has the most room to operate. The ceiling is jointly set by a construct difference between text and verdict (the cognitive mechanism) and a measurement tendency of the instrument reading the text (the negativity-bias mechanism). Both operate; the customized prompt intervenes on the second.

The aspect-level correlation pattern in Appendix B reinforces this account. The predicted rating correlates with the survey's overall rating (Spearman $\rho = 0.82$) far more strongly than with any individual aspect (Table B8: staff 0.52, concessions 0.51, entertainment 0.45, seat/view 0.43, merchandise 0.42, parking 0.36, parking exit 0.29). And the prediction delta (predicted minus survey) is essentially uncorrelated with every aspect (Table B9: $|r| \leq 0.05$ across all seven). If the under-prediction were driven by the model over-weighting a particular kind of operational friction, that aspect's rating should predict the size of the prediction error. None do. The systematic miss is not a function of which aspects fans rated lower; it is a function of how much textual space they devoted to friction relative to whatever produced their integrative verdict. The same pattern is visible at the instrument level in the empirical calibration curve: at predicted ratings of 4 through 8, the mean self-reported rating runs 0.7 to 2.2 points higher (Table B10), the calibration signature of the directional bias documented in Section 6.

This is the textual clarity ceiling: the accuracy boundary set by the relationship between what the text explicitly describes and what the survey score holistically summarizes. The ceiling is not a fixed point; it varies by the clarity of the individual text. On clear, unambiguously positive text, the ceiling is high and all capable models reach it. On mixed-signal text containing operational friction, the ceiling is substantially lower. The prompt and model operate within the space the text allows. No prompt instruction can bridge the gap in mixed-signal cases, because the information needed (how the fan weighted each element against the whole) is not in the text. It is in the number the fan chose. What prompt instruction can do is correct the measurement tendencies of the instrument reading the text—and that is the work the customized prompt is doing.

## 8. Predictions are stable across configurations, and the stability pattern mirrors the ceiling

Swap the prompt, switch the model, rerun the annotation: how much does the predicted rating change? If the answer is "substantially," the measurement is not a property of the text but an artifact of the configuration. The customized GPT 4.1 run and the generic baseline are near-identical at the session level: 79% of overlapping sessions receive exactly the same predicted rating, and 99.8% agree within ±1. GPT 4.1-mini is the outlier: it agrees exactly with the customized GPT 4.1 run on only 47% of sessions and within ±1 on 97%, the lowest cross-run agreement of any pair tested.

**Figure 7. Prediction stability across runs.**

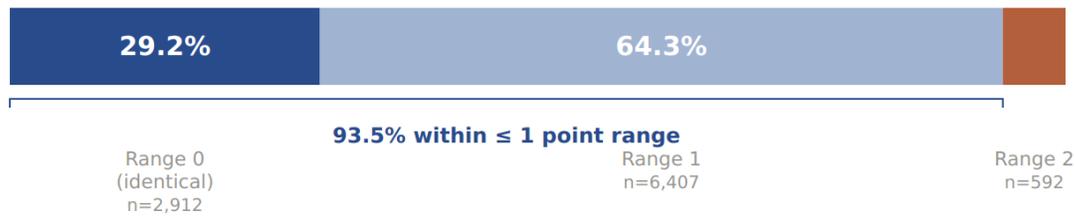

**94% of sessions fall within one point range across all four runs**
Distribution of prediction range (max − min predicted score) across all four prompt–model configurations for 9,961 sessions in the joint sample.

| 29.2% | 64.3% | |
|---|---|---|
| Range 0 (identical) n=2,912 | Range 1 n=6,407 | Range 2 n=592 |

93.5% within ≤ 1 point range

*The full bar represents all 9,961 sessions. Larger segments indicate more stable predictions across runs. The bracket below highlights the combined share within a 0–1 point range.*

prediction range across 4 runs · joint sample · n=9,961 · Figure 7

**Table 9. Pairwise session-level agreement on overlapping sessions.**

| Pair | N | Exact | Within ±1 | Mean abs. diff. |
|---|---|---|---|---|
| p2_1 vs. p3_1_41 | 9,978 | 79% | 99.8% | 0.22 |
| p2_1 vs. p3_1_52 | 9,978 | 70% | 99% | 0.31 |
| p3_1_41 vs. p3_1_52 | 9,999 | 64% | 99% | 0.37 |
| p3_1_41 vs. p3_1_41mini | 9,961 | 47% | 97% | 0.57 |

*Note. N is the per-pair bilateral overlap of valid, in-range predictions. The 9,961 for the p3_1_41mini pair (limited by 37 null predictions plus a few missing survey or transcript records) is the smallest effective N and anchors the joint sample reported in Table 2.*

The stability pattern reinforces the textual clarity ceiling from a different angle. Where all four configurations converge (93.5% of sessions vary by at most one point), the text is providing a clear enough signal that prompt and model choice are nearly irrelevant. These are the sessions where the measurement is robust to implementation decisions. Where the configurations diverge, the text is ambiguous enough that each configuration's approach to weighing surface valence against verdict produces a different answer. The practical consequence is that instability is diagnosable: the same text-rating coherence signal that predicts accuracy also predicts stability. A high-coherence prediction is not only more likely to be correct; it is more likely to be the same prediction regardless of which capable model produced it.

This relationship between text clarity and cross-configuration stability is consistent with emerging methodological work on prompt sensitivity. Barrie and Törnberg (2024) propose a Prompt Stability Score that adapts traditional intercoder reliability metrics for LLM annotation, formalizing the intuition that stable predictions signal reliable measurement. The present data extend that logic from prompt variation to model variation: when the text is clear, neither the prompt nor the model matters much. When the text is ambiguous, both matter, and the instability is itself informative about the quality of the input.

## 9. The ceiling determines what the predicted rating can and cannot do

The ceiling has a cognitive component the prompt cannot move and a measurement component the customized prompt does move. What the predicted rating can and cannot do follows from this asymmetry.

At 39% exact match, the predicted rating cannot replace the survey score as a point estimate of the fan's experience. But most operational decisions are not about point estimates. They are binary: is this fan in the top tier, or at risk? The shift from regression accuracy to classification performance is not just a different way of evaluating the same measurement. It reveals that the directional error structure documented in Sections 5 through 7, which limits the predicted rating as a substitute for the survey score, is precisely what makes it useful as a classifier.

The mechanism is straightforward. Because the model systematically under-predicts, it rarely places a dissatisfied fan in the top box. When it does assign a high rating, the text almost certainly supports that judgment. Top-box precision runs between 94% and 96% across all four configurations: when the predicted rating says a fan is in the top tier, it is right roughly 19 to 24 times out of 25. The cost is recall (61 to 77%), because some fans whose text was mixed or understated receive predicted ratings below the top-box threshold despite having rated the experience highly. The model misses them not because it read the text incorrectly but because the text did not contain enough positive signal to clear the threshold. This is the ceiling operating at the classification level. The model can only place a fan in the top box when the text unambiguously supports it, and unambiguous positive text is exactly the region where Section 7 showed accuracy is highest.

The error profile inverts for low-experience detection. Recall is high (94 to 96%): the model catches nearly every fan who self-reported a low experience, because fans who rate poorly tend to write negatively, producing texts where the signal and the verdict agree. Precision is lower (55 to 57%), because the model also flags fans whose text was more negative than their verdict. These false positives are not classification errors in the traditional sense. They are cases where the text describes a different experience than the number summarizes, the same construct divergence that produces under-prediction in the regression analysis. A fan flagged as low-experience by the model but who self-reported a moderate score is a fan whose text contained enough friction to warrant attention, even if their overall verdict was forgiving.

Table 10. Decision metrics for top-box and low-experience classification.

| Run | Top-box prec. | Top-box recall | F1 | Low-exp. prec. | Low-exp. recall | F1 |
| --- | --- | --- | --- | --- | --- | --- |
| p3_1_41 | 94% | 77% | 0.85 | 57% | 94% | 0.71 |
| p2_1 | 95% | 73% | 0.83 | 56% | 96% | 0.71 |
| p3_1_41mini | 96% | 61% | 0.75 | 55% | 96% | 0.70 |
| p3_1_52 | 94% | 74% | 0.83 | 56% | 96% | 0.71 |

One case illustrates what a false positive looks like in practice and why it may still carry operational value. A fan wrote about arriving to find their ticketed section closed, describing frustration and embarrassment with staff. The model predicted 1; the fan rated the experience a 5. This fan would appear in a low-experience outreach queue despite a moderate self-reported score. Whether that is an error depends on what the outreach queue is for. If it exists to contact fans who reported low scores, the flag is wrong. If it exists to contact fans who experienced operational friction worth addressing, the flag is right. The predicted rating and the survey score are answering different questions, and the value of the predicted rating as a classifier depends on which question the data team is asking.

The high precision of top-box classification is itself evidence that the customized prompt's bias-correction instruction is operating. A baseline LLM prone to over-weighting surface negativity would place fewer fans in the top box, because many genuinely top-tier fans write texts containing specific friction. Top-box precision above 94% means that when the model does assign a top-tier rating, it is doing so despite any surface-negative content the text contains, which is the behavior the prompt architecture is designed to produce.

## 10. Discussion

The ceiling argument has an obvious version: you cannot extract from text what is not in the text. The contribution of this paper is not the obvious version. It is a decomposition of the ceiling into components that have different causes and respond differently to engineering intervention. The ceiling on LLM-predicted experience ratings has two such components. The cognitive component is a construct difference: open-ended text records salient incidents (Schwarz 1999) while the rating scale activates an integrative judgment shaped by the peak-end heuristic and current affect (Kahneman, Wakker, and Sarin 1997). No prompt can extract from the text information the text does not contain about how the fan integrated the experience into a single number. This is the obvious half. The measurement component is an instrument bias: LLMs trained on review corpora over-weight surface negativity when mapping text to scalar ratings (Kucia 2026). This bias is in the model, not in the construct gap, and it can be intervened on. The customized prompt does exactly this, and only this. Its uniform improvement across lexical regimes, concentrated where surface negativity has the most room to mislead, is the cross-text-type signature of a bias-correction intervention rather than a general reasoning improvement. A survey-positivity-bias account is not ruled out by the data, but it predicts a uniform inflation of the survey score that the cross-text-type pattern in Section 7.2 and the aspect-correlation pattern in Section 7.3 do not show.

The decomposition explains why the engineering levers behaved as they did. Prompt customization corrected the instrument bias and gained two percentage points. Models used in this research touched neither component reliably: GPT 5.2 returned to baseline, GPT 4.1-mini fell below it. The cognitive component is invariant to both levers because it is not a property of the instrument. This indicates some potential for a stronger claim than "prompt engineering helps a little." There is a small indication of what kind of engineering helps and why. Practitioners who treat prompt design as a stylistic exercise may get noise. Practitioners who treat it as a bias-correction exercise (identify the failure modes the baseline model produces, then write instructions that specifically counter them) may realize bounded gains the instrument allows.

The decomposition speaks back to the literature it borrows from. To the LLM measurement literature, it offers a partition of where engineering effort returns and where it does not: prompt design moves the instrument-bias component, model selection does not transfer, and neither lever touches the construct component. The general injunction to validate prompt and model jointly (Carlson et al. 2026; Yang et al. 2025) becomes more specific under this account, because the validation question is not "does this configuration work?" but "which component is this configuration moving?" To the survey methodology literature, the data offer empirical evidence for a divergence Schwarz (1999) predicted but that has been hard to measure directly: open-ended text and closed-ended scales activate different cognitive processes and produce systematically different outputs from the same respondent about the same experience. The systematic, directional under-prediction documented here is what that divergence looks like at scale, in a domain where it can be quantified against ground truth. The decomposition gives both literatures a sharper version of a claim each had separately.

The decomposition also reframes the deployment categories from the prior paper (Hong et al., "Predictive Ratings," 2026). Fragile satisfaction, defined as a positive verdict resting on negatively-toned text, is the population the bias-correction lever operates on. The canonical case is a fan who rates the experience highly (e.g., a 9) but details operational friction (e.g., long concession lines or a confusing exit): a baseline negativity-biased model downgrades the prediction, but the customized prompt's decision process pulls the prediction back toward the verdict. This lever does not move the ceiling, but it sharpens the categorical separation between the two signals at the margin where the model's negativity bias was causing the misclassification. Unexpressed loyalty, the reverse case of a low rating attached to warmly-worded text, is unaffected by bias correction and remains a property of the construct gap itself, as the model has no equivalent positivity bias to correct. The decomposition specifies that both categories are part measurement artifact and part construct feature, defining which is which. The improved prompt helps the model ignore minor complaints to match high ratings, but it cannot fix mispredictions where the fan's actual reason for their score is not written in the text.

The prompt's success comes from its step-by-step instructions, but we still need to test if forcing the model to write its reasoning before assigning the score would make the correction even stronger. The bias-correction account is supported by the cross-text-type pattern but is not isolated to a single mechanism inside the customized prompt. The customized prompt bundled two interventions: a three-step decision process in the prompt body and a rationale field in the output schema. Both were generated after the rating field in the schema, which means the rationale could not condition the score it justified (Section 13). The bias correction the paper credits operates through the prompt body, which conditions all generated tokens; the rationale field as tested is a logged justification rather than a reasoning scaffold. Whether a rationale-first variant would amplify the bias-correction effect, produce an independent gain, or do nothing further is the cleanest unresolved question inside the measurement component (and is a subject for further study by the lab).

Ultimately, differences between what fans write and what they rate cannot be entirely "fixed". That is, the construct component is the larger remaining question. The text-rating gap will not close through better prompts or newer models because what is missing is not in the text. Additional inputs that supply what the text omits, and the cognitive account makes a specific prediction about what those inputs would be. If fans use residual affect as a direct informational input to the rating (Schwarz and Clore 1983), then game-level features that produce affect (outcome, competitiveness, late-inning drama) could predict the prediction delta even after controlling for text content. Whether such a test would close the cognitive component, partially explain it, or rule out the affect-as-information mechanism in favor of a different cognitive process is unknown. There are just as likely personal and emotional components that cannot be measured or accounted for with any additional data.

## 11. Conclusion

The ceiling on LLM-predicted experience ratings has two components. One is a measurement bias inside the instrument, correctable through prompt design that targets it specifically. The other is a construct difference between what fans write and what they decide, invariant to engineering on the model side because the missing information is not in the text. Prompt and model are not equivalent levers, and neither closes the construct gap. The bounded gains the customized prompt produces are exactly the gains the decomposition predicts: the measurement component moves; the cognitive component does not.

    The decomposition has a practical consequence (i.e., the limits on annotation come from the signal contained in the text itself and the ability of the LLM to align with self-reported scores when the signal does exist). The predicted rating is not simply an approximation of the survey score awaiting better engineering. It is a different measurement of the same experience, produced by a different cognitive process, with a known asymmetric error structure. The two signals are not redundant when they agree and not in conflict when they disagree. They are reading the experience from different sides. Operationally, this means the fans whose two readings disagree are the population the deployment is for: fragile satisfaction where the rating is high and the text negative, unexpressed loyalty where the reverse holds. The bias-correction lever sharpens that divergence at the margin where the instrument was misreading the text. The construct gap is what remains, and characterizing it directly is where the next investigation begins.

## 12. Implications for data teams and researchers

*The findings suggest five operational principles for teams deploying LLM-predicted ratings. Each follows from the textual clarity ceiling: the alignment of the measurement is bounded by the relationship between what the text describes and what the survey score summarizes. The most effective tactics are those that acknowledge this boundary rather than attempt to engineer past it.*

**Treat the predicted rating and the survey score as complementary signals, not substitutes**. The predicted rating captures textual salience: what the fan described. The survey score captures evaluative integration: the fan's overall verdict. These are related but structurally different constructs, and neither fully contains the other. The predicted rating is most valuable in the cases where it diverges from the survey score, because the divergence identifies fans whose written account of their experience tells a different story than their number. A data team that uses both signals can identify fans with fragile satisfaction (high rating, negative text) or unexpressed loyalty (low rating, positive text) that neither signal alone would reveal. This was the principal operational implication of previous research, and it holds without modification under the engineering interventions tested here: better prompts and different models do not close the gap between what fans write and what they report.

**Use text-rating coherence to gate prediction quality**. High-coherence predictions in the best run (p3_1_41) achieve 75% within ±1 and cover 81% of the sample (Figure 5; Table B6 reports the analogous baseline figures). Medium-coherence predictions misalign at more than twice the rate. A practical deployment strategy is to auto-accept high-coherence predictions for downstream analytics and route medium-coherence cases to human review, using the model's own assessment of how clearly the text supports its rating as the gating criterion. But coherence gating is only as useful as the model producing the labels is capable; a weak model will overestimate coherence because it cannot detect ambiguity in the text. Our previous paper established this gating pattern on the baseline prompt; the present study shows it persists under prompt customization and across the three capable model configurations, which means coherence gating is a deployable feature of the method rather than an artifact of any single prompt or model.

**Invest in prompt architecture *before* model upgrades**. Prompt customization on GPT 4.1 added roughly two percentage points of within-±1 agreement over the baseline (67% to 69%). Running the same customized prompt on GPT 5.2 returned performance to the baseline. The lever direction is asymmetric: prompt designs are shown to improve within a narrow band; model swaps, however, may not always improve performance. Prompt wrapper variation was found to have minimal effect on measurement (Asirvatham et al. 2025), but the decision architecture of the prompt, particularly instructions that explicitly counter-weight known model biases, is a separate and distinct engineering lever available. The customized prompt tested here functions as a bias-correction intervention against LLMs' documented tendency to over-weight surface negativity in scalar annotation tasks (Kucia 2026), and the uniform improvement it produces across lexical regimes is evidence that the intervention generalizes rather than being tuned to a specific text type. Teams deploying predicted ratings should treat prompt design as a bias-correction exercise, not a stylistic one: identify the failure modes the baseline model produces, then write prompt instructions that specifically counter them. Generic instructions to 'be accurate' or 'consider context' do not intervene on the mechanism causing the error. When deploying predicted ratings, optimize the prompt's reasoning structure first and validate any model change against the same ground-truth data.

**Treat textual clarity as a data quality indicator**. Within-text-type variation in agreement was more than an order of magnitude larger than variation from prompt or model choices among capable configurations, which means the input is the first place to look when predicted ratings behave poorly. Short, explicitly valenced responses produce accurate predictions; long, mixed-signal responses do not. This is the textual clarity ceiling operating at the individual session level, and it is knowable before ground truth is available. Lexical clarity buckets can serve as a front-end data quality filter alongside coherence gating.

**Monitor operational friction topics as misalignment risk factors**. Transcripts mentioning lines, wait times, parking, or exit friction are 1.3 to 1.5 times more likely to produce misaligned predictions. These topics are identifiable at annotation time using simple keyword detection. Teams can flag these cases for manual review or adjust their interpretation of the predicted rating accordingly.

**The core message about engineering and deployment**.

- *Complementary Signals: The predicted rating should be treated as a complementary signal, not a substitute for the survey score. Its operational value is highest in cases where the two diverge (identifying "fragile satisfaction" or "unexpressed loyalty").*
- *Textual Clarity is the True Ceiling: The gap in accuracy between the best and worst text types (e.g., positive-only vs. negative-only) was more than twenty times the spread between the best and worst capable configurations, confirming that the linguistic character of the input text is the binding constraint on prediction agreement.1*
- *Gating Strategy: Recommend using text-rating coherence (what the model's "confidence" label actually measures) as a reliable, pre-ground-truth way to gate prediction quality, auto-accepting high-coherence cases and routing medium-coherence cases to human review.*

## 13. Limitations

**Single domain**. This study validates predicted ratings in the context of MLB fan experience surveys. The textual clarity ceiling and the directional under-prediction pattern are consistent with findings in other domains (Kucia 2026 documents analogous under-scoring in essay evaluation), but replication across industries, survey types, and respondent populations is needed before the findings can be treated as general.

**Positively skewed sample**. Roughly half of respondents rated their experience 9 or 10. Accuracy in the moderate and low bands is estimated from smaller samples and carries wider confidence intervals. The skew is typical of post-experience satisfaction surveys, but it means the cases where the measurement is most challenged are also the cases where the statistical power is weakest.

**Survey selection bias**. Fans who complete post-game surveys are not a random sample of attendees. They are likely more engaged, more satisfied, and more willing to articulate their experience than non-respondents. This self-selection shapes both the ground-truth distribution (contributing to the positive skew) and the type of text the model encounters. Whether the ceiling operates differently on text produced by less engaged or less articulate respondents is unknown.

**No causal test of the salience-integration mechanism**. The aspect-level correlation evidence reported in Appendix B (Tables B8 and B9) is consistent with the salience-integration account but does not establish it causally. A direct test would manipulate textual salience experimentally, for example by reordering or rewriting fan responses to amplify or suppress operational details, and measure whether the predicted rating shifts in the predicted direction. This is the most natural extension of the present study.

**Single open-ended response**. The predicted rating is derived from a single text response per fan. Richer input data, whether additional survey fields, behavioral signals, or contextual metadata, might raise the accuracy ceiling by providing information the text alone does not contain.

**The design space is not exhausted**. Four configurations were tested. Other prompt strategies (few-shot examples, chain-of-thought without a structured rationale field, pairwise comparison approaches as explored by Licht et al. 2025), models, and parameter settings remain untested. The ceiling argument predicts that alternative configurations will improve performance within the space the text allows but will not raise the ceiling itself. That prediction is testable.

**The rationale field is post-hoc in both tested configurations**. In the baseline and customized prompt functions, predicted_rating is generated before any free-text field (Appendix A). Because structured-output generation is autoregressive, this means the rationale is produced conditional on a score the model has already committed to and cannot participate in the scoring decision itself. The three-step decision process in the prompt body conditions generation regardless of field order and remains the mechanism we credit for the customized prompt's bias-correction effect. But the rationale field as tested is a logged justification, not a reasoning scaffold. A variant that placed the rationale before the rating would convert it into a chain-of-thought intervention, conditioning the score on weighing logic actually executed on the specific input. Whether this changes measurement accuracy, and whether any gains would concentrate in the same lexical regimes the present bias-correction intervention helped most, is a direct next test.

# Appendix A: Prompt function schema

The analysis in this study was conducted using the Dimension Labs language data platform ([dimensionlabs.io](dimensionlabs.io)). The complete JSON schema used to generate predicted ratings is reproduced below.

```
Prompt Function for Baseline P2_1

{
  "name": "extract_session_info",
  "description": "Infer the fan's self-reported overall game-day experience
                  rating from their open-ended feedback.",
  "context": "Incoming messages are from the fan. Outgoing messages are
              survey questions.",
  "parameters": {
    "type": "object",
    "properties": {
      "predicted_rating": {
      "description": "Estimate the fan's self-reported overall experience
                      rating (0-10, 10 best) as they would answer the
                      survey question 'overall experience attending the
                      game'. Use ONLY evidence in the fan's text. If there
                      is not enough evidence to infer an overall rating
                      with confidence, output null.",
      "type": "integer"
      },
      "predicted_rating_confidence_score": {
      "description": "Confidence in the inferred overall rating based on
                      clarity and strength of evidence in the fan text.
                      High: explicit rating or very clear overall sentiment
                      with strong cues. Medium: inferable but mixed/limited
                      cues. Low: weak cues or ambiguous. If predicted_rating
                      is null, output null.",
      "type": "string",
      "enum": ["High", "Medium", "Low"]
      },
      "predicted_rating_evidence": {
      "description": "In <15 words, summarize the strongest evidence from
                      the fan text for the inferred overall rating.
                      all lower case, no punctuation. If predicted_rating
                      is null, output null.",
      "type": "string"
      }
      },
      "required": ["predicted_rating", "predicted_rating_confidence_score",
              "predicted_rating_evidence"]
  }
}
```

Prompt Function for Custom P3_1_*

```json
{
  "name": "extract_session_info",
  "description": "Infers the fan's self-reported overall game-day experience
                  rating from their open-ended feedback.",
  "context": "Incoming messages are from the fan. Outgoing messages are survey
              questions.",
  "parameters": {
    "type": "object",
    "properties": {
      "predicted_rating": {
      "description": "Estimate the fan's self-reported overall experience
                      rating (0-10) as they would answer the survey
                      question.\n\nDecision process:\n1. If the comment
                      contains an explicit overall verdict or rating, treat
                      the 'overall' verdict as primary.\n2. Otherwise, infer
                      the *verdict* by weighing both positives and negatives
                      in the text.\n3. Do not over-penalize a minor complaint
                      if the comment is otherwise strongly positive.",
        "type": "integer"
      },
      "predicted_rating_rationale": {
      "description": "Explain the rationale for the score assigned. all lower
                      case, no punctuation. If predicted_rating is null, output
                      null.",
      "type": "string"
      },
      "predicted_rating_confidence_score": {
      "description": "Confidence in the inferred overall rating based on
                      clarity and strength of evidence in the fan text. High:
                      explicit rating or very clear overall sentiment with
                      strong cues. Medium: inferable but mixed/limited cues.
                      Low: weak cues or ambiguous. If predicted_rating is null,
                      output null.",
        "type": "string",
        "enum": ["High","Medium","Low"]
      },
      "predicted_rating_evidence": {
      "description": "In <15 words, summarize the strongest evidence from the
                      fan text for the inferred overall rating. all lower
                      case, no punctuation. If predicted_rating is null,
                      output null.",
      "type": "string"
      }
    },
    "required": ["predicted_rating", "predicted_rating_rationale",
        "predicted_rating_confidence_score", "predicted_rating_evidence"]
  }
}
```

# Appendix B: Supplementary Tables

### Table B1. Performance by survey rating level, all four configurations

*Section 7 argues that the alignment ceiling is concentrated in the midrange of the rating scale, where survey verdicts must reconcile mixed evidence the model treats as a discount; this table makes that concentration visible by reporting agreement metrics at every integer point of the survey scale.*

| Configuration | Survey rating | n | Exact match | Within ±1 | MAE |
|---|---|---|---|---|---|
| p2_1 | 0 | 234 | 12.8% | 34.2% | 1.86 |
| p2_1 | 1 | 185 | 17.8% | 68.1% | 1.34 |
| p2_1 | 2 | 250 | 36.4% | 82.8% | 0.88 |
| p2_1 | 3 | 268 | 39.6% | 85.1% | 0.81 |
| p2_1 | 4 | 267 | 24.0% | 66.3% | 1.17 |
| p2_1 | 5 | 603 | 12.4% | 39.6% | 1.72 |
| p2_1 | 6 | 541 | 12.4% | 34.4% | 2.11 |
| p2_1 | 7 | 1,144 | 23.3% | 45.5% | 2.01 |
| p2_1 | 8 | 1,751 | 26.8% | 57.7% | 1.79 |
| p2_1 | 9 | 1,548 | 29.9% | 69.3% | 1.45 |
| p2_1 | 10 | 3,187 | 61.7% | 89.5% | 0.52 |
| p3_1_41 | 0 | 236 | 11.9% | 31.8% | 2.00 |
| p3_1_41 | 1 | 186 | 15.1% | 62.4% | 1.45 |
| p3_1_41 | 2 | 250 | 37.2% | 80.4% | 0.91 |
| p3_1_41 | 3 | 268 | 36.9% | 85.4% | 0.84 |
| p3_1_41 | 4 | 268 | 24.3% | 68.3% | 1.13 |
| p3_1_41 | 5 | 607 | 13.5% | 41.7% | 1.67 |
| p3_1_41 | 6 | 542 | 12.0% | 34.3% | 2.05 |
| p3_1_41 | 7 | 1,145 | 22.6% | 47.0% | 1.95 |
| p3_1_41 | 8 | 1,752 | 29.7% | 60.3% | 1.66 |
| p3_1_41 | 9 | 1,554 | 34.0% | 72.5% | 1.33 |
| p3_1_41 | 10 | 3,191 | 67.9% | 93.1% | 0.42 |
| p3_1_41mini | 0 | 235 | 3.4% | 9.8% | 2.44 |
| p3_1_41mini | 1 | 185 | 4.9% | 43.2% | 1.71 |
| p3_1_41mini | 2 | 249 | 32.9% | 83.9% | 0.92 |
| p3_1_41mini | 3 | 267 | 53.6% | 91.8% | 0.60 |
| p3_1_41mini | 4 | 268 | 25.0% | 76.9% | 1.01 |
| p3_1_41mini | 5 | 606 | 12.7% | 40.8% | 1.58 |
| p3_1_41mini | 6 | 541 | 12.2% | 31.1% | 2.06 |
| p3_1_41mini | 7 | 1,141 | 22.9% | 44.1% | 1.99 |
| p3_1_41mini | 8 | 1,750 | 25.9% | 56.7% | 1.79 |
| p3_1_41mini | 9 | 1,548 | 27.8% | 64.8% | 1.52 |
| p3_1_41mini | 10 | 3,171 | 12.0% | 77.0% | 1.16 |
| p3_1_52 | 0 | 236 | 11.0% | 44.9% | 1.67 |
| p3_1_52 | 1 | 186 | 33.3% | 80.1% | 1.03 |
| p3_1_52 | 2 | 250 | 45.6% | 87.6% | 0.74 |
| p3_1_52 | 3 | 268 | 26.5% | 80.2% | 1.02 |
| p3_1_52 | 4 | 268 | 16.0% | 53.0% | 1.41 |
| p3_1_52 | 5 | 607 | 6.8% | 32.0% | 2.04 |
| p3_1_52 | 6 | 542 | 14.9% | 29.7% | 2.34 |
| p3_1_52 | 7 | 1,146 | 21.6% | 46.9% | 2.16 |
| p3_1_52 | 8 | 1,752 | 24.0% | 58.3% | 1.85 |
| p3_1_52 | 9 | 1,554 | 35.3% | 69.4% | 1.40 |
| p3_1_52 | 10 | 3,191 | 49.6% | 90.5% | 0.63 |

**Table B2. Error directionality across configurations**

*The central interpretive claim of the paper—that misalignment is dominated by under-prediction rather than symmetric noise—rests on the consistently negative mean signed error and the asymmetry between over- and under-scored shares reported here for all four configurations.*

| Configuration | n | Mean signed error | % over | % under | % exact | % |err|≥2 | % |err|≥3 |
|---|---|---|---|---|---|---|---|
| p2_1 | 9,978 | -0.97 | 10.9% | 52.7% | 36.4% | 32.9% | 18.3% |
| p3_1_41 | 9,999 | -0.83 | 12.6% | 48.0% | 39.4% | 30.6% | 17.0% |
| p3_1_41mini | 9,961 | -1.18 | 9.9% | 70.2% | 19.9% | 38.6% | 18.6% |
| p3_1_52 | 10,000 | -1.09 | 10.3% | 57.4% | 32.4% | 32.9% | 19.5% |

**Table B3. Confidence gating: coverage and performance by self-reported confidence**

*Section 12 recommends routing low-confidence predictions to human review or follow-up; this table quantifies the operational tradeoff that recommendation rests on by showing the within-±1 gap between High and Medium strata for each configuration, together with the share of sessions each stratum covers.*

| Configuration | Confidence | n | Coverage | Exact match | Within ±1 | MAE |
|---|---|---|---|---|---|---|
| p2_1 | High | 5,866 | 58.8% | 50.7% | 81.4% | 0.879 |
| p2_1 | Medium | 4,112 | 41.2% | 16.0% | 46.8% | 1.894 |
| p3_1_41 | High | 8,108 | 81.1% | 45.3% | 75.1% | 1.065 |
| p3_1_41 | Medium | 1,891 | 18.9% | 14.1% | 44.5% | 1.863 |
| p3_1_41mini | High | 8,909 | 89.4% | 21.0% | 64.6% | 1.437 |
| p3_1_41mini | Medium | 1,052 | 10.6% | 9.9% | 34.4% | 2.128 |
| p3_1_52 | High | 4,668 | 46.7% | 49.2% | 76.9% | 1.069 |
| p3_1_52 | Medium | 5,330 | 53.3% | 17.7% | 58.5% | 1.659 |

**Table B4. Distribution stability: pairwise distances between predicted-rating histograms**

*Section 8 frames distributional stability as a property of the prompt-and-model configuration rather than of the underlying construct; the pairwise total variation distances reported here show that p3_1_41mini is a distributional outlier while the other three configurations remain close to one another.*

| Configuration A | Configuration B | L1 distance | Total variation distance |
| --- | --- | --- | --- |
| p2_1 | p3_1_41 | 0.090 | 0.045 |
| p2_1 | p3_1_52 | 0.200 | 0.100 |
| p3_1_41 | p3_1_52 | 0.263 | 0.131 |
| p2_1 | p3_1_41mini | 0.416 | 0.208 |
| p3_1_41 | p3_1_41mini | 0.485 | 0.243 |
| p3_1_41mini | p3_1_52 | 0.458 | 0.229 |

## Table B5. Performance by transcript length

*Reviewers commonly ask whether richer text improves inferred ratings; this robustness check shows that within each configuration the relationship between transcript length and agreement is essentially flat or mildly negative, indicating that the alignment ceiling is not a function of how much the fan wrote.*

| Configuration | Transcript length (chars) | n | Exact match | Within ±1 | MAE |
|---|---|---|---|---|---|
| p2_1 | <500 | 8,002 | 37.7% | 68.4% | 1.262 |
| p2_1 | 500–999 | 1,682 | 31.3% | 62.3% | 1.448 |
| p2_1 | 1000–1999 | 256 | 30.9% | 59.8% | 1.410 |
| p2_1 | 2000–3999 | 30 | 26.7% | 63.3% | 1.333 |
| p2_1 | 4000+ | 5 | 20.0% | 40.0% | 1.600 |
| p3_1_41 | <500 | 8,021 | 41.3% | 71.0% | 1.168 |
| p3_1_41 | 500–999 | 1,684 | 32.1% | 63.0% | 1.404 |
| p3_1_41 | 1000–1999 | 256 | 28.5% | 59.8% | 1.422 |
| p3_1_41 | 2000–3999 | 30 | 23.3% | 60.0% | 1.500 |
| p3_1_41 | 4000+ | 5 | 40.0% | 40.0% | 1.400 |
| p3_1_41mini | <500 | 7,985 | 19.0% | 62.4% | 1.504 |
| p3_1_41mini | 500–999 | 1,682 | 22.7% | 57.7% | 1.550 |
| p3_1_41mini | 1000–1999 | 256 | 28.1% | 57.0% | 1.418 |
| p3_1_41mini | 2000–3999 | 30 | 20.0% | 46.7% | 1.667 |
| p3_1_41mini | 4000+ | 5 | 20.0% | 40.0% | 1.600 |
| p3_1_52 | <500 | 8,022 | 33.1% | 68.7% | 1.339 |
| p3_1_52 | 500–999 | 1,684 | 29.3% | 61.1% | 1.569 |
| p3_1_52 | 1000–1999 | 256 | 30.5% | 57.0% | 1.535 |
| p3_1_52 | 2000–3999 | 30 | 30.0% | 63.3% | 1.500 |
| p3_1_52 | 4000+ | 5 | 20.0% | 60.0% | 1.800 |